\pdfoutput=1
\documentclass[11pt]{article}

\usepackage{EMNLP2022}
\usepackage{multirow}
\usepackage{times}
\usepackage{latexsym}
\usepackage{braket}
\usepackage{amsfonts}
\usepackage[inline]{enumitem}

\usepackage{graphicx} %插入图片的宏包
\usepackage{float} %设置图片浮动位置的宏包
\usepackage{subfigure} %插入多图时用子图显示的宏包
\usepackage[ruled,linesnumbered]{algorithm2e}
\usepackage{algorithmic}

\newcommand{\mb}{\mathbf}

\usepackage{tikz}
\usepackage{pgfplots}
\usepackage{pgfplotstable}
\usepackage{makecell}
\usepackage{microtype}
\usepackage{graphicx}
\usepackage{subfigure}
\usepackage{booktabs} % for professional tables

\usepackage{times}
\usepackage{latexsym}
\usepackage{amsthm}

\usepackage[T1]{fontenc}
\usepackage[utf8]{inputenc}
\usepackage{amsmath}

\newtheorem{theorem}{Theorem}

\usepackage{microtype}

\usepackage{inconsolata}

\title{RAPO: An Adaptive Ranking Paradigm for Bilingual Lexicon Induction}

\author{Zhoujin Tian\thanks{\quad Work is done during internship at Microsoft.}, Chaozhuo Li\thanks{\quad Corresponding author and equal contribution.}, Shuo Ren, Zhiqiang Zuo, Zengxuan Wen, Xinyue Hu \\
        {\bf Xiao Han, Haizhen Huang, Denvy Deng, Qi Zhang, Xing Xie} \\
        Microsoft \\ \texttt{deritt7@gmail.com}, \texttt{\{cli,renshuo,zhiqzuo,zewen,xinyuehu\}@microsoft.com} \\
        \texttt{\{xiaoha,hhuang,dedeng,qizhang,xingx\}@microsoft.com}}

\begin{document}
\maketitle
\begin{abstract}

Bilingual lexicon induction induces the word translations by aligning independently trained word embeddings in two languages. 
Existing approaches generally focus on minimizing the distances between words in the aligned pairs,  while suffering from low discriminative capability to distinguish the relative orders between positive and negative candidates. 
In addition, the mapping function is globally shared by all words, whose performance might be hindered by the deviations in the distributions of different languages. 
In this work, we propose a novel ranking-oriented induction model RAPO to learn personalized mapping function for each word.  
RAPO is capable of enjoying the merits from the unique characteristics of a single word and the cross-language isomorphism simultaneously.
Extensive experimental results on public datasets including both rich-resource and low-resource languages demonstrate the superiority of our proposal. Our code is publicly available in \url{https://github.com/Jlfj345wf/RAPO}.

\end{abstract}

\section{Introduction}

Bilingual lexicon induction (BLI) aims at inducing the word translations across two languages based on the monolingual corpora, which is capable of transferring valuable semantic knowledge between different languages, spawning a myriad of NLP tasks such as  machine translation \citep{lample-etal-2018-phrase, artetxe2018unsupervised}, semantic parsing \citep{xiao-guo-2014-distributed}, and document classification \citep{klementiev-etal-2012-inducing}. 
The nucleus of BLI is learning a desirable mapping function to align two sets of independently trained monolingual word embeddings \citep{DBLP:journals/corr/MikolovLS13, DBLP:journals/jair/RuderVS19, glavas-etal-2019-properly}.  
\citet{DBLP:journals/corr/MikolovLS13} empirically observe that the linear projections are superior to their non-linear counterparts due to the isomorphism across different embedding spaces. 
Subsequent improvements are successively proposed to advance BLI task by imposing orthogonal constraints \citep{xing-etal-2015-normalized, DBLP:conf/iclr/LampleCRDJ18}, normalizing the embeddings \citep{artetxe-etal-2016-learning, DBLP:conf/aaai/ArtetxeLA18, zhang-etal-2019-girls}, reducing the noises \citep{artetxe-etal-2018-robust, yehezkel-lubin-etal-2019-aligning}, relaxing the hypothesis of isomorphism \citep{sogaard-etal-2018-limitations, patra-etal-2019-bilingual}, and iteratively refining the seed dictionary \cite{zhao-etal-2020-semi}. 

\begin{figure}
	\centering
	\includegraphics[width=\linewidth]{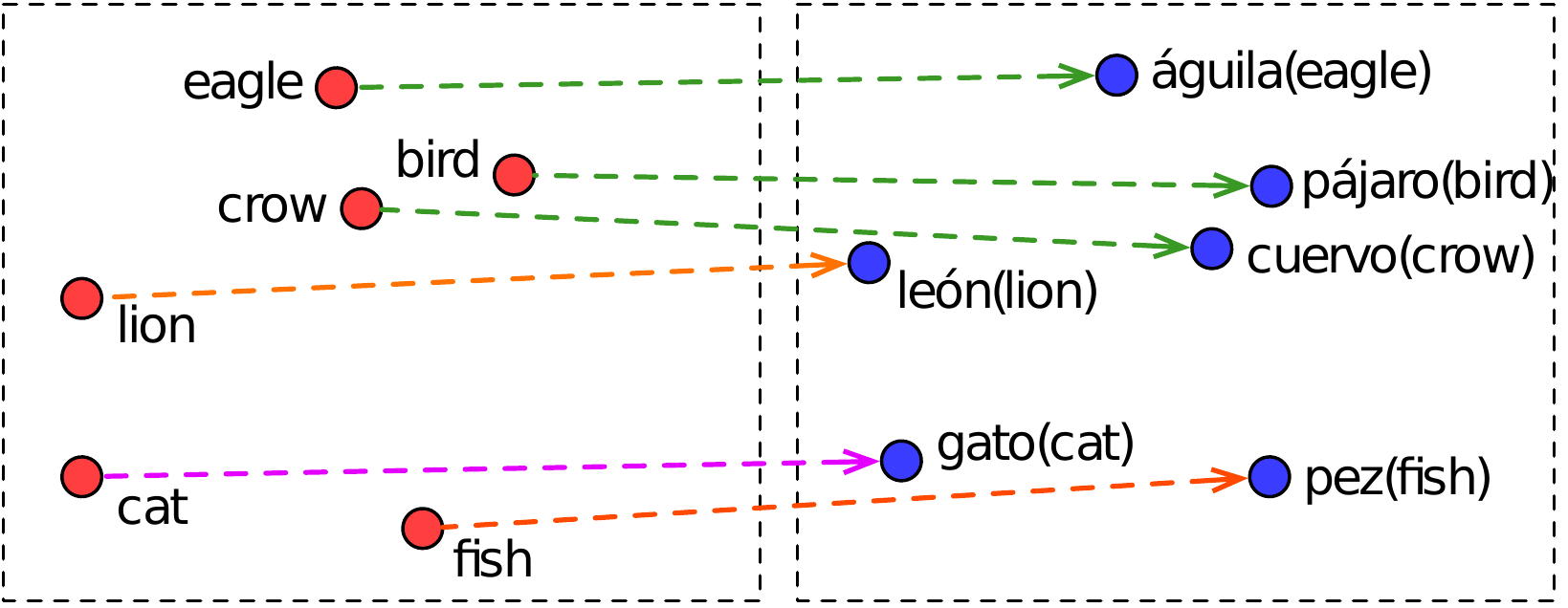}
	\caption{Representation spaces of several aligned words in English (left) and Spanish (right).}
	\label{fig:nonlinear}
\end{figure}

Existing methods \citep{artetxe-etal-2016-learning, DBLP:conf/aaai/ArtetxeLA18, jawanpuria-etal-2020-geometry} usually aim at minimizing the distance between the word from the source language and its aligned word in the target language (e.g., \textit{crow} and \textit{cuervo} in Figure \ref{fig:nonlinear}).    
However, BLI is essentially a ranking-oriented task because for each source word, we expect to select its top $k$ high-confidence target candidates. 
Namely, a desirable BLI model should also be capable of distinguishing the relative orders between the positive and negative candidates (e.g., \textit{crow} and \textit{curevo} should be distributed closer than \textit{crow} and \textit{pájaro}). 
The objective functions used by previous works solely focus on the distances between positive pairs and cannot explicitly provide such important ranking signals, leading to the low discriminative capability.  

% The discriminative capability of induction models could be further strengthened by explicitly  exploiting such ranking signals, which has great potential to advance model performance. 

% Previous works only focus on the correlations between words in the aligned pairs, but cannot distinguish the relative orders between the positive and negative candidates, resulting in the discrepancy and sub-optimal performance. 

% For example, conventional approaches  

% Generally, BLI is essentially a ranking task because for each source word, we expect to select top $k$ high-confidence candidates from the target language.  
% Hence, a desirable mapping function should satisfy the following two prerequisite:  \textbf{(i)} minimize the distance between word $s$ from the source language and its aligned target word $t$;   
% \textbf{(ii)} ensure the distance between positive pair $\braket{s,t}$ is less than the defective pairs $\braket{s,\hat{t}}$. 
% However, existing works usually learn the mapping function via minimizing the distances between the aligned seeds, which only focus on the prerequisite \textbf{(i)} but ignore \textbf{(ii)}. 
% Such methods might be incapable of distinguishing the relative orders between the positive and negative pairs, resulting in the discrepancy and sub-optimal performance. 
% Therefore, how to design a ranking-based training paradigm is essential to the BLI task. 

In addition, conventional BLI models \citep{DBLP:journals/corr/MikolovLS13, xing-etal-2015-normalized, zhao-etal-2020-semi} induce the bilingual space via a shared mapping function, in which different words in the same language tend to be rotated in the same directions. 
However, several studies \citep{sogaard-etal-2018-limitations, patra-etal-2019-bilingual} have demonstrated that the isomorphic assumption may not strictly hold true in general, and thus a global-shared mapping is not the optimal solution.  
As shown in Figure \ref{fig:nonlinear}, even for two close languages like English and Spanish, due to deviations in the distributions of different training corpora and insufficient training of low-frequency word embeddings, the optimal mapping directions are slightly shifted for different words. 
Therefore, the BLI performance could be further improved if we could learn unique or personalized mapping functions for different words.  
\citet{glavas-vulic-2020-non} first propose to achieve the personalized mappings. 
However, \citet{glavas-vulic-2020-non} is a non-parametric model, in which the personalized mappings are unlearnable and built upon the  heuristic assumptions, which might be unreliable and suffer from low generality.

%我们提出了一种新的学习框架, 消除了训练和测试之间的gap,并且为每个单词提供了个性化的调整。
To address the mentioned limitations under a unified framework, we propose a novel \textbf{R}anking-based model with \textbf{A}daptive \textbf{P}ersonalized \textbf{O}ffsets, dubbed \textbf{RAPO}. 
Different from previous works solely relying on the aligned pairs, RAPO is formulated as a ranking paradigm with powerful discriminative capability by incorporating abundant unaligned negative samples. 
An effective dynamic negative sampling strategy is further proposed to optimize the ranking objectives. 
Moreover, we integrate a novel personalized adapter into RAPO to learn unique mapping directions for different words. 
A straightforward strategy is to directly learn an independent mapping matrix for each word, which is resource-consuming and ignores the global isomorphism information. 
Differently, our personalized adapter learns the unique offset for each word to calibrate the vanilla embedding, and then a shared mapping function is employed to induct lexicons. 
By organically integrating personalized offsets with shared mapping functions, RAPO enjoys the merits from the unique traits of each word and the global consistency across languages.   
We further propose a novel Householder projection as the mapping function on the basis of Householder matrices \citep{householder1958unitary}, which strictly ensures the orthogonality during the model optimization. 
% Personalized adapter calibrates the vanilla word embeddings under the guidance from the induction losses. 
% calibrate the vanilla representation spaces. 
% The embedding adapter can be trained through gradient descent and is capable of learning the personalized offsets based on the contextual semantic information, which provides unique rotation directions for different words.  
We conduct extensive experiments over multiple  language pairs in the public MUSE benchmarks \citep{DBLP:conf/iclr/LampleCRDJ18}, including rich- and low-resource languages. 
Experimental results demonstrate that our proposal consistently achieves desirable performance in both supervised and semi-supervised learning settings. 
%Our code is publicly available in  \url{https://github.com/lixin4ever/Conference-Acceptance-Rate}. 
% Our code is publicly available in \url{https://github.com/Jlfj345wf/RA-BLI}.

Our contributions are summarized as follows:
\begin{itemize}
	\item To the best of our knowledge, we are the first to propose a ranking-based bilingual lexicon induction model RAPO with powerful discriminative capacity.   
	
	\item We propose a novel personalized adapter to achieve unique mapping direction for each word by adaptively learning the personalized embedding offsets. 
	
	\item We conduct extensive experiments over popular benchmarks and the results demonstrate the superiority of our proposal. 
\end{itemize}

\section{Preliminary}

Let $\mb{X} \in \mathbb{R} ^ {d \times n_x}$ and $\mb{Y} \in \mathbb{R} ^ {d \times n_y}$ be monolingual embedding matrices consisting of $n_x$ and $n_y$ words for the source and target languages respectively,  and $d$ stands for the embedding size. $\mathcal{D} = \{(\mb{x}_1, \mb{y}_1), ..., (\mb{x}_l, \mb{y}_l)\}$ denotes the available aligned seeds, which can be also formulated as two matrices $\mathbf{X}_D$ and $\mathbf{Y}_D \in \mathbb{R}^{d \times l}$. 
BLI aims to map the source and target words from their original embedding spaces into a shared latent space, in which the mapped source word $\phi_s(\mb{x}_i)$ should be close to its matched target word $\phi_t(\mb{y}_i)$. 
$\phi_s$ and $\phi_t$ denote the mapping functions for source language and target language, respectively. For the sake of clarification, notations used in this paper are listed in Appendix \ref{sec:notations}.  
 
A widely adopted solution is to set the source mapping function $\phi_s$ as a linear transformation matrix $\mathbf{W} \in \mathbb{R}^{d \times d}$ and the target mapping function $\phi_t$ to an identity matrix $\mathbf{I} \in \mathbb{R}^{d \times d}$.  The objective function is defined as follows:  
\begin{equation}
	\label{eq:baseline_obj}
		\mathbf{W^{*}}= \underset{\mathbf{W}}{\arg \min } \ \|\mathbf{W}\mathbf{X}_D -\mathbf{I}\mathbf{Y}_D\|_{F}.  
\end{equation} 
% Based on the orthogonal Procrustes solvers \citep{artetxe-etal-2016-learning},  a closed-form solution obtained from singular value decomposition (SVD) of $\mathbf{Y X}^{\top}$ as follows:
% \begin{equation}
% 	\label{eq:svd_map}
% 		\mathbf{W}^{*}=\mathbf{U V}^{\top} 
% 	 \text { with } \mathbf{U} \Sigma \mathbf{V}^{\top}=\operatorname{SVD}\left(\mathbf{Y}\mathbf{X}^{\top}\right). 
% \end{equation} 
In the inference phase, the distances between the mapped source words and the target ones are utilized as the metrics to select the top-$k$ nearest neighbors as the translations. 

\begin{figure*}
	\centering
	
	\includegraphics[width=0.9\linewidth]{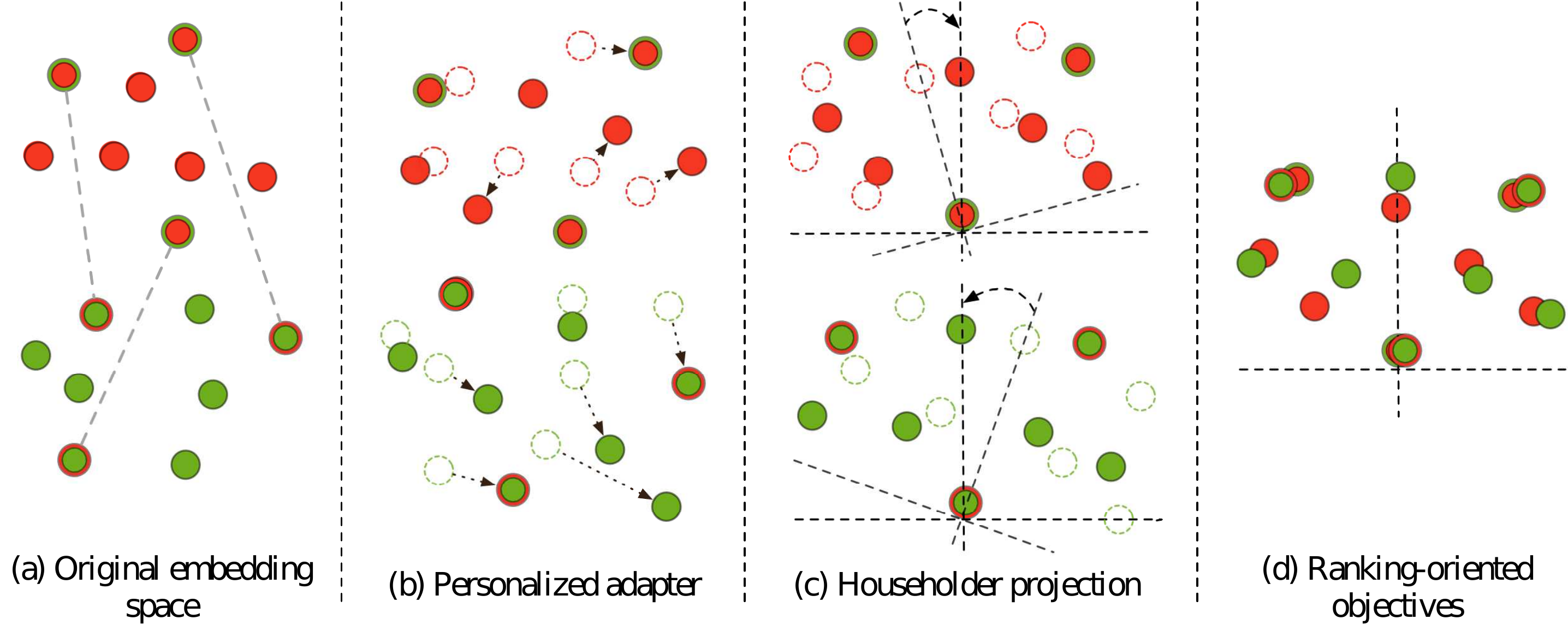}
	\caption{Overview of the proposed RAPO model. First, the personalized adapter calibrates vanilla embeddings based on the learned personalized offsets. Then, two adjusted embedding spaces are mapped into a shared latent space via the orthogonal Householder projections. Finally, we design the hybrid ranking-oriented objective functions to optimize the model parameters.}
	\label{fig:model}
\end{figure*}

As discussed in the section of introduction, this popular induction paradigm suffers from two limitations. 
First, Formula (\ref{eq:baseline_obj}) only focuses on minimizing the distance between the aligned words (e.g., $\mb{x}_{i}$ and $\mb{y}_{i}$). 
However, BLI task is essentially a ranking problem, which means that the learned mapping function should also be able to distinguish the aligned pair \{$\mb{x}_{i}$,  $\mb{y}_{i}$\} and the defective one \{$\mb{x}_{i}$,  $\mb{y}_{j}$\}. 
Second, the globally shared mapping matrix $\mathbf{W}$ might be inappropriate since the optimal mapping directions of different words tend to be various.  
Our proposed RAPO is capable of addressing the mentioned challenges under a unified learning framework. 
We will introduce the details of RAPO in the next section.

\section{Methodology}

% 先讲一下问题定义
% 说一下我们的framework如图所示 总结一下 分三部分
% 第一部分 adapter
% 第二部分 householder
% 第三部分 negative 
% 第四部分 learning object 
% 第五部分 procedure & algorithm code

As shown in Figure \ref{fig:model}, RAPO consists of three major components. 
Given the monolingual embeddings and training seeds, the personalized adapter first generates the adaptive offset for each word by exploiting the contextual semantic information. 
The vanilla embedding spaces are properly calibrated to fit the induction task. 
After that, we map the adapted word embeddings to a shared latent space via the novel Householder projections, which is capable of ensuring the strict orthogonality and better preserving the isomorphism. 
Finally, RAPO designs the ranking objectives to distinguish the aligned pairs from the unmatched ones. 
RAPO can be easily adapted to the supervised and semi-supervised settings, demonstrating its flexibility.   
 
\subsection{Personalized adapter}

%Previous work empirically showed that the word embedding spaces of different languages exhibit similar geometric structures. Although recent work found that the isomorphic assumption does not necessarily hold true, they still proposed to use an orthogonal or approximate orthogonal matrix as the mapping matrix, because a fully non-linear model inevitably suffer from overfitting on small training dictionary. Therefore, we apply the language special orthogonal transformation to map the adjusted monolingual embeddings into a common latent space respectively. In which case, the adjusted embeddings ensure the robustness of the model to non-isomorphic scenarios, and the orthogonal transformation ensures the generalization of the model to the totally unseen test data.

%Recent studies have empirically showed that the isomorphic assumption does not hold true in general and providing the same mapping for each word in the entire embedding space leads to a sub-optimal results \citep{sogaard-etal-2018-limitations, patra-etal-2019-bilingual}. Due to deviations in the distribution of different training corpora and insufficient training of low-frequency word embeddings, each word should be globally projected with some personalised offset which may be very different from each other to align with the target word more accurately. \citep{glavas-vulic-2020-non} proposes a post-processing method that allows each word to have a special rotation, but their pre-defined rules may be unreliable and lack robustness.

Due to the deviations in the distribution of different corpora and the unbalanced training of word embeddings, recent works demonstrated that the vanilla word embedding space may not be fully trustworthy and proper adjustments contribute to improving induction performance.      
Previous work \citep{glavas-vulic-2020-non} proposes to modify the mapped embedding based on its nearest neighbors in the
training dictionary, which is a non-parametric model and might be unreliable. 
Here we design a novel learnable personalized adaptor, which can be trained through the gradient descent and learn task-relevant personalized offsets.   

% generates the personalized offsets based on the contextual semantic information, which provides unique rotation directions for different words.  

% which can be trained through gradient descent and is capable of learning the personalized offsets based on the contextual semantic information, which provides unique rotation directions for different words.  

% Despite its promising performance,  INSTAMAP \citep{glavas-vulic-2020-non} is a non-parametric model.  Its embedding rectification solely relies on the heuristic assumptions, which might be inconsistent with the learning signals. 
% Different from previous non-parametric method \citep{glavas-vulic-2020-non}, we propose a novel personalized embedding adapter to calibrate the original embeddings by generating learnable offsets.
% %inspired by FiLM(\citep{DBLP:conf/aaai/PerezSVDC18}). 

Given a source word embedding $\mb{x}$, adapter first obtains its contextual semantic vector $\bar{\mb{x}}$ by averaging the embeddings of its neighbor words $\mathcal{M}_s(\mb{x})$ in the source space. 
Our motivation lies in that a single word can only provide limited information, while a set clustering similar words can assemble the mutual word relationships and provide richer and more accurate information. 
The contextual semantic vector $\bar{\mb{x}}$ is formally defined as: 
\begin{equation}
    \label{csc}
	\begin{aligned}
		\mb{\bar{x}} ~ =& ~ \frac{1}{m_s} \sum_{\mb{x}_j \in \mathcal{M}_s(\mb{x})}\mb{x}_j \\
		\mathcal{M}_s(\mb{x})~=&~\{\mb{x}_j \ | \ \braket{\mb{x}_j, \mb{x}} > \tau_s\}
	\end{aligned}
\end{equation} 
where $m_s$ is the size of $\mathcal{M}_s(\mb{x})$ and  $\braket{,}$ denotes the dot product. 
$\tau_s$ is a hyper-parameter denoting the similarity threshold. 
Compared to the original word embedding $\mb{x}$, the contextual vector $\mb{\bar{x}}$ is more informative by incorporating richer semantics. 

% A set clustering similar words usually has similary meaning with each other
% this is doubtful because the isomorphic assumption
% may not hold all the time (Søgaard et al., 2018).
% Fortunately, the cliques we extracted naturally provide good local features for us, because they are usually much different from each other in meanings, which enables us to investigate alternatives to a single mapping matrix W.

After that, personalized adapter learns the unique offset for each word based on the contextual semantic vector, which can be effectively optimized by the training objectives. 
Previous work \citep{ren-etal-2020-graph} has observed that semantic similar words enjoy stronger isomorphism structures across different embedding spaces. Thus, our motivation lies in that words with similar contextual semantics also tend to have similar personalized offsets.

Specifically, the adapter is implemented as a feed-forward network with a single layer:
\begin{equation}
    \label{ada}
	\begin{aligned}
		\operatorname{A}_s(\mb{x}) = \sigma(\mb{W}_{s}\mb{\bar{x}}) 
	\end{aligned}
\end{equation}
where $\sigma$ denotes the activation function and it could be linear or non-linear. $\mb{W}_s \in \mathbb{R} ^ {d \times d}$ stands for the learnable parameters. 
The generated offset vector indicates the personalized offset direction, and is further combined with the vanilla embedding $\mb{x}$: 
\begin{equation}
\label{adapter_calc}
	\begin{aligned}
		\tilde{\mb{x}} = \mb{x} + \operatorname{A}_s(\mb{x}) = \mb{x} + \sigma(\mb{W}_{s}\mb{\bar{x}}).
	\end{aligned}
\end{equation}
 $\tilde{\mb{x}}$ denotes the calibrated word embedding, and will be normalized to the unit length to ensure the consistent value range. 

Similarly, for the target word embedding $\mb{y}$,  the calibrated embedding $\tilde{\mb{y}}$ can be calculated as:
\begin{equation}
\label{adapter_calc_y}
	\begin{aligned}
		\tilde{\mb{y}} = \mb{y} + \operatorname{A}_t(\mb{y}) = \mb{y} + \sigma(\mb{W}_{t}\mb{\bar{y}})
	\end{aligned}
\end{equation}
where $\operatorname{A}_t(\mb{y})$ is the personalized adapter for target language with learnable parameters $\mb{W}_{t} \in \mathbb{R} ^ {d \times d}$ and the contextual semantic vector $\bar{\mb{y}}$ is obtained in the similar manner.

In a nutshell, the proposed adapter has the following obvious advantages. 1) \textbf{Personalization}: the offset is learned based on the contextual semantic features, which are different for various words.  2)  \textbf{Flexibility}: $\sigma$ could be either linear or non-linear function to handle different types of language pairs such as close languages (e.g., English-Spanish) or distant ones (e.g., English-Chinese). 
3) \textbf{Task-relevant}: vanilla word embeddings are unsupervised learned and might be incompatible with the BLI task. The proposed adapter is capable of properly adjusting the original embeddings according to the downstream induction tasks.

\subsection{Householder projection}

Based on the calibrated embeddings, we further need to design desirable mapping functions to map them into a shared latent space. 
Previous works \citep{xing-etal-2015-normalized, DBLP:conf/iclr/LampleCRDJ18, patra-etal-2019-bilingual} have demonstrated that the orthogonality of the mapping function is crucial to the model performance.  
% previous works(Patra-et-al-2019,Mohiuddin-et-al-2020) usually add additional restrictions to the mapping matrix , which can only 
% achieve approximated orthogonal transformations but not strict ones.
A general approach is to add an extra constraint in the objective function to force the mapping matrix to be orthogonal (i.e., min$_{W}$  $||\mathbf{W}\mathbf{W}^\top-\mathbf{I}||_2$). Nevertheless, such constraints can only achieve an approximate orthogonal matrix instead of the strict one, which may hinder the capability of BLI models in capturing the unsupervised isomorphism information.
Here we propose to construct a strict orthogonal mapping function based on the Householder matrices \citep{householder1958unitary,li2022house}, dubbed Householder projection. 

% 斜体的H？ 
Householder matrix represents the reflection about a hyperplane containing the origin. 
Given a unit vector $\mb{v} \in \mathbb{R}^d$, the $d \times d$ Householder matrix $\mb{H}$, taking $\mb{v}$ as parameter, is defined as $\operatorname{H}(\mb{v})$:
\begin{equation}
	\begin{aligned}
		\operatorname{H}(\mb{v}) = \mb{I} - 2\mb{v}\mb{v}^{\top}
	\end{aligned}
\end{equation}
where $||\mb{v}||_2=1$ and $\mb{I}$ is the $d \times d$ identity matrix.
Given an input vector $\mathbf{z}$, the Householder matrix transforms $\mb{z}$ to $\mb{\hat{z}}$ by a reflection about the hyperplane orthogonal to the normal vector $\mb{v}$:
\begin{equation}
\label{HT_vec}
	\begin{aligned}
		\mb{\hat{z}} = \operatorname{H}(\mb{v})\mb{z} = \mb{z} - 2\braket{\mb{z}, \mb{v}}\mb{v}.
	\end{aligned}
\end{equation}

Based on the Householder matrix, we can design a novel Householder projection as the mapping function to ensure strict orthogonal transformations. 
Householder projection is composed of a set of consecutive Householder matrices. 
Specifically, given a series of unit vectors $\mathcal{V} =  \{\mb{v}_{i}\}_{i=1}^{n}$ where $\mb{v}_i\in \mathbb{R}^d$ and $n$ is a positive integer, we define the Householder Projection (HP) as follows:
\begin{equation}
	% \vspace{-1mm}
	\begin{split}
		\label{map-Hrot}
		{\rm HP}(\mathcal{V}) = \prod_{i=1}^{n}\operatorname{H}(\mb{v}_{i}).
	\end{split}
	% \vspace{-1mm}
\end{equation} 
We can theoretically prove the following theorem: 
\begin{theorem}
\label{thm-1}
% Any $k \times k$ orthogonal matrix $Q$ can be decomposed into the product of $k-1$ or $k$ Householder matrices.
The image of ${\rm HP}$ is the set of all $n \times n$ orthogonal matrices, i.e., ${\rm Image(HP)}=\boldsymbol{{\rm O}}(n)$, $\boldsymbol{{\rm O}}(n)$ is the $n$-dimensional orthogonal group. (See proof in Appendix \ref{proof})
\end{theorem} 

% Specifically, given a series of vectors $U =  \{u_i\}_{i=1}^{n}$ where $u_i\in \mathbb{R}^k$ and $n$ is a positive integer, we define the mapping as follows:
% \begin{equation}
% 	% \vspace{-1mm}
% 	\begin{split}
% 		\label{map-Hrot}
% 		{\rm Orth}(U) = \prod_{i=1}^{n}H(u_i).
% 	\end{split}
% 	% \vspace{-1mm}
% \end{equation} 

% We can prove that the image of ${\rm Orth}$ is the set of all $n \times n$ orthogonal matrices, i.e., ${\rm Image(Orth)}=\boldsymbol{{\rm O}}(n)$, $\boldsymbol{{\rm O}}(n)$ is the $n$-dimensional orthogonal group. 

Next we will introduce how to employ the Householder projections in the RAPO model. 
Each language is associated with its unique Householder projection to map words into the shared latent space. 
Take the source language as an example. 
Given the calibrated source word embedding $\tilde{\mb{x}}$, we employ the source  Householder projection with $\mathcal{V}_s = \{\mb{v}_1,..,\mb{v}_n\}$ as the mapping function:
\begin{equation}
\label{HT}
\begin{aligned}
\mb{\hat{x}} &= \operatorname{HP}(\mathcal{V}_s)\tilde{\mb{x}} = \prod \limits_{i=1}^n \operatorname{H}(\mb{v}_i)\tilde{\mb{x}} 
\end{aligned}
\end{equation}
where $n$ is set to $d$ to fully cover the set of all $d \times d$ orthogonal matrices. Similarly, we employ $n$ unit vectors $\mathcal{U}_t = \{\mb{u}_1,..,\mb{u}_n\}$ to parameterize the Householder projection for the target language. 

Based on Theorem \ref{thm-1}, the Householder projection is capable of maintaining strict orthogonality during model optimization such as stochastic gradient descent (SGD), which is theoretically appealing compared to its counterparts. 
In addition, the efficiency  of Householder projection is also guaranteed. 
The number of learnable parameters in Householder projection (i.e., the set $\mathcal{V}_{s}$) is $d \times d$, which is identical to the conventional mapping matrix $\mb{W}$ in Formula (\ref{eq:baseline_obj}). 
Moreover, the matrix-vector multiplications in Formula (\ref{HT}) can be replaced by vector multiplications following Formula (\ref{HT_vec}), which reduce the time complexity from $O(nd^2)$ to $O(nd)$.

\subsection{Ranking-oriented objective function}
BLI is a ranking-oriented task as we expect to select the top $k$ target words with high confidence for each source word. 
However, the training objective functions (e.g., Formula (\ref{eq:baseline_obj})) of previous works \citep{artetxe-etal-2016-learning, DBLP:conf/aaai/ArtetxeLA18, jawanpuria-etal-2020-geometry} essentially minimize the distances between the aligned words, which cannot provide sufficient discriminative capacity to rank the candidates.  
Differently, we propose to optimize the RAPO model by a ranking loss, which empowers our proposal the ability of learning the relative orders between the aligned words and unmatched ones. 

Specifically, we adopt a popular pair-wise ranking loss,  Bayesian personalized ranking (BPR) \citep{DBLP:conf/uai/RendleFGS09,zhang2022geometric}, as the the major training objective function. 
Given an aligned positive pair $(\mb{x},\mb{y})$, the ranking loss is defined as follows: 
\begin{equation}
	\begin{aligned}
		&\mathcal{L}_{r}(\mb{x},\mb{y}) = \\
		&- \frac{1}{K} \sum_{\mb{\hat{y}}^{-} \in \mathcal{N}^{-}}\log\delta(g(\mb{\hat{x}},\mb{\hat{y}}) - g(\mb{\hat{x}}, \mb{\hat{y}}^{-})) 
	\end{aligned}
\end{equation} 
in which $\delta$ is the sigmoid function, and g(,) measures the similarity between two word embeddings. 
In order to counteract the challenge of hubness problem, we adopt the cross-domain similarity local scaling (CSLS) \citep{joulin-etal-2018-loss} as the similarity metric, which penalizes the similarity values in dense areas of the embedding distribution. 

Set $\mathcal{N}^{-}$ contains $K$ negative samples, which provides crucial ranking signals for RAPO. Most of the negative sampling strategies can be divided into hard negative sampling and random negative sampling. Following the theoretical analysis of these two strategies \citep{DBLP:conf/sigir/ZhanM0G0M21}, we propose to utilize dynamic hard negatives to optimize the top-ranking performance and random negatives to stabilize the training procedure. 

Hard negative sampling emphasizes the top-ranking performance and disregards the lower-ranked pairs that hardly affect the lexicon inductions. 
We employ the dynamic hard negative sampling strategy as the traditional static sampling methods have potential risks in decreasing ranking performance \citep{DBLP:conf/sigir/ZhanM0G0M21}. 
% We aim to select the top-ranked negative candidates in each training epoch.
At the beginning of each training epoch, we retrieve the top-$K_{h}$ candidates for each source word based on the current model parameters, from which the unmatched samples are selected as the hard negative samples.  
To stabilize the training process, we additionally introduce random negative samples. For each source word in training dictionary, $K_{r}$ random negative samples are uniformly sampled from the entire target vocabulary. 
Finally, we can achieve $K = K_h + K_r$ negative samples by merging the hard and random samples. 

Additionally, another loss is incorporated to emphasize on the supervised signals by minimizing the Euclidean distance between the aligned words: 
\begin{equation}
	\begin{aligned}
		\mathcal{L}_{m}(\mb{x}, \mb{y}) = ||\mb{\hat{x}} - \mb{\hat{y}} ||_2
	\end{aligned}
\end{equation}

The final loss of RAPO is the combination of these two objective functions: 
\begin{equation}
\label{loss}
	\begin{aligned}
		\mathcal{L}(\mathcal{D}, \theta) =& \frac{1}{l}\sum_{(\mb{x},\mb{y}) \in \mathcal{D}} (\mathcal{L}_{r}(\mb{x},\mb{y}) + \lambda_1 \mathcal{L}_{m}(\mb{x},\mb{y})) \\ 
		& + \lambda_2 ||\theta||_2 
	\end{aligned}
\end{equation}
where $\theta$ denotes the model parameter set, $\lambda_1, \lambda_2$ are hyper-parameters to control the importance of  losses and the last item is the L2 regularization to control the range  of parameter values.

\begin{algorithm}[t]
    \small
    \caption{Training procedure of RAPO}
    \label{procedure}
    \begin{algorithmic}[1] % 此处的[1]控制一下算法中的每句前面都有标号 
        \REQUIRE Monolingual word embeddings $\mb{X}$ and $\mb{Y}$, training dictionary $\mathcal{D}$, number of iterations $C$, learning rate $\alpha$, number of epochs $n\_epochs$. 
        \STATE Pre-process contextual semantic vectors (Formula (\ref{csc})).  
        \FOR {$c \leftarrow 1$ \KwTo $C$ }
            \FOR {$e \leftarrow 1$ \KwTo $n\_epochs$ }
                % 这里用一个sample来表示合适吗
                \STATE {Generate the negative sample set $\mathcal{{N}^{-}}$ including both random and dynamic hard negative samples.} 
                \STATE {Achieve personalized offsets with Formula (\ref{ada}).}
                \STATE {Calibrate source and target embeddings following Formula (\ref{adapter_calc}) and (\ref{adapter_calc_y}), respectively.}
                \STATE {Map the source and target words using the Householder projection with Formula (\ref{HT}).}
                \STATE {Calculate the loss $\mathcal{L}$ with Formula (\ref{loss}).}
                % 这里用一个sample来表示合适吗
                
                %\STATE {$\mathcal{L}_{rank}(\mb{x}_i,\mb{y}_i) \leftarrow$} 
                %\STATE {$- \frac{1}{K} \sum_{\mathcal{\hat{N}}(\mb{x}_i)}\log\sigma(r(\mb{x}'_i,\mb{y}'_i) - r(\mb{x}'_i, \hat{\mb{y}}'_{ij}))$}
                %\STATE {$\mathcal{L}_{mse}(\mb{x}_i,\mb{y}_i) \leftarrow || \mb{x}'_i - \mb{y}'_i ||_2$}
                \STATE {$\theta \leftarrow$ parameters of $\{\mb{W}_s, \mb{W}_t, \mathcal{V}_s, \mathcal{U}_t\}$}
                %\STATE {$\mathcal{L}(\mathcal{D}, \theta) \leftarrow$}
                %\STATE {$\frac{1}{N}\sum_{ \mathcal{D}} \mathcal{L}_{rank} + \lambda_1 \frac{1}{N}\sum_{ \mathcal{D}}\mathcal{L}_{mse} + \lambda_2 ||\theta||_2 $}
                \STATE {$\theta \leftarrow \theta - \alpha \partial \mathcal{L}(\mathcal{D}, \theta) $}

            \ENDFOR
            \STATE {/* dictionary augmentation */}
            \STATE {Calculate mapped embeddings: $\mb{\hat{X}}, \mb{\hat{Y}}$}. 
            \STATE {Induce new seeds $\mathcal{D}'$ based on the CSLS similarities between $\mb{\hat{X}}$ and $\mb{\hat{Y}}$.}
            \STATE {$\mathcal{D} \leftarrow  \mathcal{D} \cup \mathcal{D}'$}
        \ENDFOR
    \end{algorithmic} 
\end{algorithm}

\subsection{Training paradigm} 

The proposed RAPO model can be employed in both supervised and semi-supervised  scenarios. 
Here we introduce the semi-supervised training procedure of RAPO  in  Algorithm \ref{procedure}, which iteratively expands the seed dictionary and refines the mapping function. 
We first calculate the training loss, and optimize the parameters of the adapters and householder projections as shown in line 3-11. 
After that,  we augment the seed dictionary by selecting the mutual nearest CSLS neighbors after each training iteration in line 13-15. 
In the inference phase, both source and target words are mapped into the shared latent space, and the nearest CSLS neighbor of each source word are selected as its translation. We also conduct complexity analysis in Appendix \ref{sec:Complexity}.

\section{Experiment}

\subsection{Experimental Settings}

\paragraph{Dataset}   RAPO model is evaluated on the the widely used MUSE dataset \citep{DBLP:conf/iclr/LampleCRDJ18}. 
Following \cite{mohiuddin-etal-2020-lnmap}, we select five high-resource language pairs and five low-resource pairs to thoroughly test the model performance. 
Precision@1 is selected as the measurement \cite{DBLP:conf/iclr/LampleCRDJ18}. 
We use the data splits in the original MUSE dataset. Detailed statistics could be found in Appendix \ref{sec:datasets}.

\paragraph{Baselines}  
We compare RAPO with popular SOTA BLI baselines, including unsupervised methods \cite{DBLP:conf/iclr/LampleCRDJ18, artetxe-etal-2018-robust, mohiuddin-joty-2019-revisiting, ren-etal-2020-graph} and semi-supervised/supervised methods \cite{DBLP:conf/aaai/ArtetxeLA18, DBLP:conf/iclr/LampleCRDJ18, joulin-etal-2018-loss, jawanpuria-etal-2019-learning, patra-etal-2019-bilingual, mohiuddin-etal-2020-lnmap, zhao-etal-2020-semi}. 
Please refer to Appendix \ref{relatedwork} for the details. 
For each baseline, we directly report the results in the original papers and conduct experiments with the publicly available code if necessary. 

\paragraph{Implementation details}   
Following previous works, vocabularies of each language are trimmed to the 200K most frequent words.  
The original word embeddings are normalized to enhance performance \cite{artetxe-etal-2018-robust}, including a length normalization, center normalization and another length normalization. 
The number of training iterations is set to 5, and the number of training epochs is set to 150 with early stopping. 
We use the CSLS as the induction metric, and the number of nearest neighbors in the CSLS is set to 10. 
Adam optimizer is selected to minimize training loss. 
We only consider 15,000 most frequent words in the seed dictionary augmentation \cite{zhao-etal-2020-semi}. 
The search spaces of hyper-parameters are presented in Appendix \ref{sec:hypter_search}.

\subsection{Main Results}
The proposed RAPO model is extensively evaluated over five rich-resource language pairs (en-es, en-fr, en-it, en-ru and en-zh) and five low-resource pairs (en-fa, en-tr, en-he, en-ar and en-et) in both directions, leading to 20 evaluation sets in total.  
RAPO model is fully trained 5 times over each dataset and we report the average performance. Table \ref{tab:museresult} and \ref{tab:museresult_lowresource} present the Precision@1 scores of different models. 

From the results, one can clearly see that the proposed RAPO model achieves best performance over  most datasets, and obtain comparable results on other datasets. 
From the perspective of average performance, RAPO-sup outperforms the best baseline by 0.7\% and 1.3\% over the rich- and low-resource datasets, and  RAPO-semi beats the best baseline by 0.7\% and 1.6\% under the semi-supervised setting. 
Such consistent performance gains demonstrate the superiority and generality of RAPO. 
Besides, RAPO achieves more significant improvements over the distant language pairs (e.g., en-ru, en-fa and en-tr), which indicates that the personalized adapter is capable of bridging the gaps between language pairs by calibrating the vanilla embeddings to fit the BLI task.   
%\czl{By enjoying the merits of self-learning, the performance of RAPO-semi is superior that its supervised version, which reveals the effectiveness of the proposed semi-supervised learning paradigm. }
Overall, such advanced performance of RAPO owes to the appropriate ranking objectives, personalized adaptions and orthogonal projections.

\begin{table*}[t!]
\small
  \centering
  %\resizebox{1\columnwidth}{!}{
  \setlength{\tabcolsep}{2mm}{
    \begin{tabular}{l|cc|cc|cc|cc|cc|c}
      \hline
     \multirow{2}{*}{Method} & \multicolumn{2}{c|}{en-es}  & \multicolumn{2}{c|}{en-fr} & \multicolumn{2}{c|}{en-it}  & \multicolumn{2}{c|}{en-ru}  & \multicolumn{2}{c|}{en-zh} & avg \\
       &$\rightarrow$ &  $\leftarrow$ &$\rightarrow$ & $\leftarrow$ &$\rightarrow$ & $\leftarrow$ &$\rightarrow$ & $\leftarrow$ &$\rightarrow$ & $\leftarrow$ & \\
      \hline
      \multicolumn{12}{c}{Unsupervised}  \\
      \hline
      \cite{DBLP:conf/iclr/LampleCRDJ18}       &81.7 &83.3  &82.3 &81.1 &77.4 &76.1 &44.0 &59.1 &32.5 &31.4 & 64.9\\
      \cite{artetxe-etal-2018-robust}    & 82.3 &84.7   & 82.3 &83.6 &78.8 &79.5 &49.2 &65.6 &- &-&- \\
      \cite{mohiuddin-joty-2019-revisiting}  & 82.7 &84.7   & - & - & 79.0 &79.6 &46.9&64.7&-&-& - \\
      \cite{ren-etal-2020-graph} & 82.9 &85.3  &82.9 &83.9  &79.1 &79.9 &49.7 &64.7  &38.9 &35.9 & 68.3\\   
      \hline
      \multicolumn{12}{c}{Supervised}                                   \\
      \hline
      \cite{DBLP:conf/aaai/ArtetxeLA18}   & 81.9 & 83.4  & 82.1 &82.4 &77.4 &77.9 &51.7 &63.7 &32.3 &43.4 & 67.6\\
      \cite{joulin-etal-2018-loss}  & \underline{84.1}  & \underline{86.3}   & 83.3 & 84.1 & 79.0 &80.7 &57.9 &67.2 &45.9 & \underline{46.4} & 71.6\\
      \cite{jawanpuria-etal-2019-learning}  & 81.9 &85.5 & 82.1 &84.2 & 77.8 & 80.9  &52.8 & 67.6 & 49.1 &45.3 & 70.7\\
      \hline
      \textbf{RAPO}-sup  & \underline{84.1} & 86.1  & \underline{83.5} & \underline{84.3} & \underline{79.3} & \underline{81.9}  & \underline{58.1} & \underline{68.0} & \underline{51.7} & 45.9 & 72.3 \\
      \hline
      \multicolumn{12}{c}{Semi-Supervised}                                   \\
      \hline
      \cite{patra-etal-2019-bilingual}     &84.3 &86.2  & 83.9 &84.7 &79.3 &82.4 & 57.1&  67.7& 47.3 & 46.7 & 72.0 \\      
      \cite{mohiuddin-etal-2020-lnmap}     &80.5 &82.2  &- &- &76.7 &78.3 & 53.5& 67.1 & - & -& -\\   
      \cite{zhao-etal-2020-semi}$_{CSS}$  &  84.5 &86.9 & \textbf{85.3} &85.3 & \textbf{81.2} &82.7 & 57.3 & 67.9& 48.2 & 47.1 & 72.6 \\   
      \cite{zhao-etal-2020-semi}$_{PSS}$     &83.7 &86.5 &84.4 & 85.5 &80.4 &82.8 & 56.8&  67.4& 48.4 & 47.5 & 72.3\\  
      \cite{glavas-vulic-2020-non}     &82.4&	86.3& 84.5	&	84.9&	80.2&	81.9&	57.0&	67.1&	47.9&	47.2&	71.9\\   
      % \bottomrule
      \hline
      \textbf{RAPO}-semi    &\textbf{84.5} & \textbf{87.0} & 85.0 & \textbf{85.7} & 80.8  & \textbf{83.1} & \textbf{59.4} & \textbf{68.2} & \textbf{51.9} & \textbf{47.7} & \textbf{73.3}\\
      \hline
    %   \multicolumn{13}{c}{Semi-Supervised use "all 5k"}      \\
    %   \hline
    %   BLISS(R) & 84.3 & - & 79.1 & - & 83.9 & - & 79.3   &  -   & 57.1 & 67.7 & 48.7 & 47.3 \\
    %   LNMAP(linear/non-linear) \\
    %   CSS/PSS RCSLS  \\
    %   \hline
    \end{tabular}
  }
  \caption{Precision@1 for the BLI task on five rich-resource language pairs.  Best results are in \textbf{bold} and the best supervised results are \underline{underlined}. The improvements are statistically significant (sign test, p-value < 0.01)}.
  \label{tab:museresult}
\end{table*}

\begin{table*}[t!]
\small
  \centering
  %\resizebox{1\columnwidth}{!}{
  \setlength{\tabcolsep}{2mm}{
    \begin{tabular}{l|cc|cc|cc|cc|cc|c}
      \hline
     \multirow{2}{*}{Method} & \multicolumn{2}{c|}{en-fa}  & \multicolumn{2}{c|}{en-tr} & \multicolumn{2}{c|}{en-he}  & \multicolumn{2}{c|}{en-ar}  & \multicolumn{2}{c|}{en-et} & avg \\
       &$\rightarrow$ &  $\leftarrow$ &$\rightarrow$ & $\leftarrow$ &$\rightarrow$ & $\leftarrow$ &$\rightarrow$ & $\leftarrow$ &$\rightarrow$ & $\leftarrow$ & \\
      \hline
      \multicolumn{12}{c}{Unsupervised}  \\
      \hline
      \cite{DBLP:conf/iclr/LampleCRDJ18}       &33.4&	40.7&	52.7&	63.5&	43.8&	57.5&	33.2&	52.8&	33.7&	51.2&	46.2\\
      \cite{artetxe-etal-2018-robust}    & 30.5	&-&	46.4&	-&	36.8&	53.1&	29.3&	47.6&	19.4&	-&	- \\
      \cite{mohiuddin-joty-2019-revisiting}  & 36.7	&44.5&	51.3&	61.7&	44.0&	57.1&	36.3&	52.6&	31.8&	48.8&	46.5 \\
      %\cite{ren-etal-2020-graph} & 82.9 &85.3  &82.9 &83.9  &79.1 &79.9 &49.7 &64.7  &38.9 &35.9 &\\   
      \hline
      \multicolumn{12}{c}{Supervised}                                   \\
      \hline
      \cite{DBLP:conf/aaai/ArtetxeLA18}   & 39.0&	42.6&	52.2&	63.7&	47.6&	58.0&	41.2&	\underline{55.5}&	37.4&	54.0&	49.1\\
      \cite{joulin-etal-2018-loss}  & 40.5&	42.4&	53.8&	61.7&	52.2&	57.9&	42.2&	\underline{55.5} &	40.0&	50.2&	49.6\\
      \cite{jawanpuria-etal-2019-learning}  & 38.0&	40.9&	48.6&	61.9&	43.1&	56.7&	38.1&	53.3&	33.7&	48.7&	46.3\\
      \hline
      \textbf{RAPO}-sup  & \underline{41.0} & \underline{43.9}  & \underline{54.2} & \underline{64.1} & \underline{53.5} & \underline{58.5}  & \underline{43.0} & 55.3 & \underline{40.7} & \underline{55.3} & \underline{50.9} \\
      \hline
      \multicolumn{12}{c}{Semi-Supervised}                                   \\
      \hline
      \cite{patra-etal-2019-bilingual}     &38.4&	39.3&	51.8&	59.6&	51.6&	55.2&	41.1&	53.9&	36.3&	48.3&	47.6  \\      
      \cite{mohiuddin-etal-2020-lnmap}     &36.8 &	43.7&	52.5&	65.3&	52.5&	\textbf{59.1}&	42.2&	57.1&	41.2&	57.5&	50.8 \\   
      \cite{zhao-etal-2020-semi}$_{CSS}$    &41.4&	45.8&	53.1&	63.8&	53.0&	57.8&	44.1&	\textbf{57.2}&	39.2&	49.4&	50.5 \\   
      \cite{zhao-etal-2020-semi}$_{PSS}$     &41.8&	46.1&	54.0&	65.4&	49.8&	57.4&	40.2&	55.5&	40.9&	50.8&	50.2\\ 
      \cite{glavas-vulic-2020-non}     &40.3&	45.2&	54.3&	64.5&	47.5&	56.6&	41.6&	56.4&	40.1&	52.7&	50.1\\   
      % \bottomrule
      \hline
      \textbf{RAPO}-semi    & \textbf{42.4} & \textbf{46.3} & \textbf{55.7}&	\textbf{65.8}&	\textbf{53.5}&	58.7&	\textbf{44.8}&	56.5&	\textbf{42.5}&	\textbf{58.1} &	\textbf{52.4}\\
      \hline
    %   \multicolumn{13}{c}{Semi-Supervised use "all 5k"}      \\
    %   \hline
    %   BLISS(R) & 84.3 & - & 79.1 & - & 83.9 & - & 79.3   &  -   & 57.1 & 67.7 & 48.7 & 47.3 \\
    %   LNMAP(linear/non-linear) \\
    %   CSS/PSS RCSLS  \\
    %   \hline
    \end{tabular}
  }
  \caption{Precision@1 for the BLI task on five low-resource language pairs.  Best results are in \textbf{bold} and the best supervised results are \underline{underlined}. The improvements are statistically significant (sign test, p-value < 0.01)}
  \label{tab:museresult_lowresource}
\end{table*}

% \paragraph{Results in semi-supervised setting}

% As shown in Table \ref{tab:museresult} and \ref{tab:museresult_lowresource}, we compare our method with previous supervised / semi-supervised methods on resource-rich language pairs and low-resource languages pairs.
% We find that our proposed method significantly outperforms previous methods on most tasks and has comparable results on others, especially on low-resource languages and distant languages, with the improvements of 1 to 3 points compared with previous state-of-the-art approaches. The improvement in the average results on these language pairs shows that our method has a stable performance on various language pairs compared with other methods.

% \paragraph{Comparing with supervised method without dictionary expandation}
% To better demonstrate the superiority of our method in aligning word pairs, we compare our method with previous supervised / semi-supervised methods with a fixed training dictionary on resource-rich language pairs, which means we remove the expansion of the dictionary during training iterations in all of these methods.
% As shown in Table \ref{tab:museresult}, We find that our proposed method also outperforms previous methods on most tasks with the improvements of 1 to 2 points. The results on some tasks such as en-zh and ru-en are remarkably competitive with strong semi-supervised methods, which shows that the learning ability of our model is better than other methods.

\begin{table}[t!]
  \centering
  \resizebox{1\columnwidth}{!}{
    \begin{tabular}{l|cc|cc|cc|cc}
      \hline
     \multirow{2}{*}{Models} & \multicolumn{2}{c|}{en-it}  & \multicolumn{2}{c|}{en-ru} & \multicolumn{2}{c|}{en-tr} & \multicolumn{2}{c}{en-he} \\
       &$\rightarrow$ &  $\leftarrow$ &$\rightarrow$ & $\leftarrow$ &$\rightarrow$ & $\leftarrow$ &$\rightarrow$ & $\leftarrow$ \\
      \hline 
      \textbf{Best} & \textbf{79.3} & \textbf{81.9} & \textbf{58.1} & \textbf{68.0} & \textbf{54.2} & \textbf{64.1} & \textbf{53.5}&	\textbf{58.5} \\ 
       \hline
      w/o PA & 78.9 & 80.8 & 57.6 & 66.9  & 53.4 & 63.5 & 52.8 & 57.9 \\
     \hline
     linear & \textbf{79.3} & \textbf{81.9} & 57.9 & 67.4 & 53.8 &   63.9 & 53.0 & 58.1   \\
     tanh   & 78.9 & 80.9 & 57.8 & \textbf{68.0} & \textbf{54.2} &   \textbf{64.1} & 53.2 & \textbf{58.5}   \\
     sigmoid & 79.0 & 81.3 & \textbf{58.1} & 67.6 & 53.7 &   63.8 & \textbf{53.5} & 58.4 \\ 
      \hline
    \end{tabular}
  }
  \caption{Ablation study on the personalized adaptor.}
  \label{tab:ab_pa}
\end{table}

\subsection{Ablation Study}

To investigate the effectiveness of various components in RAPO, we conduct extensive ablation studies on four datasets, including en-it, en-ru, en-tr and en-he. 
To avoid the influence of the potential noises introduced by the self-learning, ablation studies are investigated under the supervised setting.

\paragraph{Personalized adaptor} Here we study the impact of the personalized adaptor (PA) module. 
Table \ref{tab:ab_pa} reports the experimental results. 
After removing the personalized adaptor, model performance consistently drops over all the datasets, revealing the importance of personalized embedding adaption. 
Furthermore, we also investigate the influence of activation function $\sigma$ in Formula (\ref{adapter_calc}) and (\ref{adapter_calc_y}). 
From Table \ref{tab:ab_pa}, we can observe that the linear activation function works better for the closer language pairs (e.g., en-it), while the non-linear functions are more suit for the distant pairs. This observation is aligned with previous works \citep{mohiuddin-etal-2020-lnmap}. 
Thus, the activation function $\sigma$ should be carefully tuned based on the data characteristics.

\begin{table}[t!]
  \centering
  \resizebox{1\columnwidth}{!}{
    \begin{tabular}{l|cc|cc|cc|cc}
      \hline
     \multirow{2}{*}{Models} & \multicolumn{2}{c|}{en-it}  & \multicolumn{2}{c|}{en-ru} & \multicolumn{2}{c|}{en-tr} & \multicolumn{2}{c}{en-he} \\
       &$\rightarrow$ &  $\leftarrow$ &$\rightarrow$ & $\leftarrow$ &$\rightarrow$ & $\leftarrow$ &$\rightarrow$ & $\leftarrow$ \\
      \hline 
      \textbf{Best} & \textbf{79.3} & \textbf{81.9} & \textbf{58.1} & \textbf{68.0} & \textbf{54.2} & \textbf{64.1} & \textbf{53.5}&	\textbf{58.5} \\ 
       \hline
      w/o HA & 78.4 & 81.0 & 57.2 & 67.2  & 52.9 & 63.2 & 52.7 & 57.2 \\
     constraint & 78.9 & 81.2 & 57.6 & 67.7 & 53.6 &   63.7 & 53.2 & 57.9   \\
      \hline
    \end{tabular}
  }
  \caption{Ablation study on the Householder projection.}
  \label{tab:ab_hp}
\end{table}

\paragraph{Householder projection} Here we aim to study the importance of Householder projections (HP), and results are reported in Table \ref{tab:ab_hp}.  
One can see that all the performance declines without the Householder projections. 
After employing the conventional orthogonal constraint $\mathbf{W}\mathbf{W}^\top = \mathbf{I}$, the performance is improved but still inferior to the Householder projections. 
We can get the following two conclusions: 1) the orthogonal mapping functions can boost the BLI performance; 2) the Householder projections are capable of ensuring strict orthogonality and thus obtaining better results. 

\begin{table}[t!]
  \centering
  \resizebox{1\columnwidth}{!}{
    \begin{tabular}{l|cc|cc|cc|cc}
      \hline
     \multirow{2}{*}{Models} & \multicolumn{2}{c|}{en-it}  & \multicolumn{2}{c|}{en-ru} & \multicolumn{2}{c|}{en-tr} & \multicolumn{2}{c}{en-he} \\
       &$\rightarrow$ &  $\leftarrow$ &$\rightarrow$ & $\leftarrow$ &$\rightarrow$ & $\leftarrow$ &$\rightarrow$ & $\leftarrow$ \\
      \hline 
      \textbf{Best} & \textbf{79.3} & \textbf{81.9} & \textbf{58.1} & \textbf{68.0} & \textbf{54.2} & \textbf{64.1} & \textbf{53.5}&	\textbf{58.5} \\ 
       \hline
      w/o $\mathcal{L}_{r}$ & 78.3 & 80.3 & 56.9 & 67.0  & 53.1 & 63.0 & 52.2 & 57.4 \\
     w/o $\mathcal{L}_{m}$ & 79.0 & 81.4 & 57.7 & 67.7 & 53.5 &  63.5 & 52.9 & 58.0   \\
      \hline
    \end{tabular}
  }
  \caption{Ablation study on the objective functions.}
  \label{tab:ab_rof}
\end{table}

\paragraph{Training objective functions} 
As shown in Formula (\ref{loss}), the loss of RAPO includes two parts: the ranking loss $\mathcal{L}_{r}$ and the MSE loss $\mathcal{L}_{m}$. Table \ref{tab:ab_rof} presents the model performance without different objectives. 
We can see that both models present performance decay over all the datasets, which verifies both objective functions would benefit the BLI task. 
Without the ranking loss, RAPO presents more significant performance drop compared to the MSE loss. 
It reveals that the BLI task is essentially a ranking problem and thus the ranking loss $\mathcal{L}_{r}$ would be more important. 

\subsection{Parameter sensitivity analysis}
The parameter sensitivity analysis is  conducted on four core hyper-parameters: the number of hard negative samples $K_{h}$, the number of random negative samples $K_{r}$, the number of Householder matrices (HM) $n$ in Formula (\ref{map-Hrot}) and the adaptor threshold $\tau_{s}$ in Formula (\ref{csc}).  
The sensitivity analysis is conducted on the en$\rightarrow$zh and en$\rightarrow$tr datasets under the semi-supervised learning scenario. 

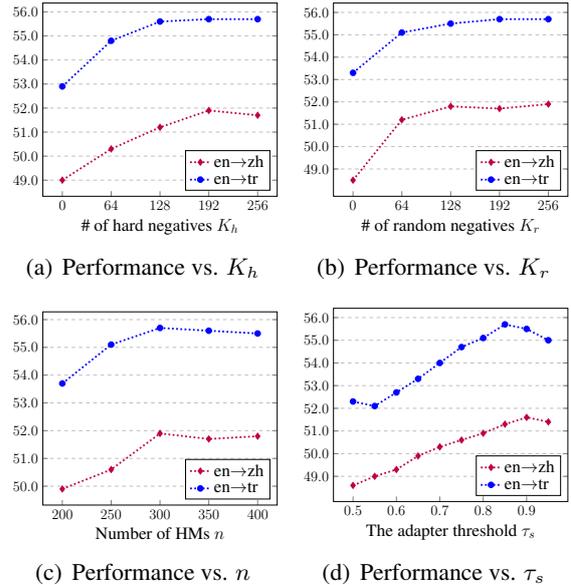
\begin{figure}[t!]
\centering
\label{fig:parameters}
    \subfigure[Performance vs. $K_{h}$]{\label{fig:parameters_kh}
            \begin{tikzpicture}[font=\Large, scale=0.45]
                \begin{axis}[
                    legend cell align={left},
                    legend style={nodes={scale=1.0, transform shape}},
                    xlabel={\# of hard negatives $K_h$},
                    xtick pos=left,
                    tick label style={font=\large},
                    %ylabel style={font=\large},
                    %xmode=log,
                    %log ticks with fixed point,
                    %ylabel={p1},
                    %xmin=0.005, xmax=100.0,
                    %ymin=0.33, ymax=0.6,
                    xtick={ 0, 64, 128, 192, 256},
                    xticklabels={ $0$, $64$, $128$, $192$, $256$},
                    ytick={49.0, 50.0, 51.0,52.0,53.0,54.0,55.0,56.0},
                    yticklabels={$49.0$, $50.0$, $51.0$, $52.0$, $53.0$, $54.0$, $55.0$, $56.0$},
                    legend pos=south east,
                    ymajorgrids=true,
                    grid style=dashed
                ]
                \addplot[
                    color=purple,
                    dotted,
                    mark options={solid},
                    mark=diamond*,
                    line width=1.5pt,
                    mark size=2pt
                    ]
                    coordinates {
                    (0, 49.0)
                    (64, 50.3)
                    (128, 51.2)
                    (192, 51.9)
                    (256, 51.7)
                    };
                    \addlegendentry{en$\rightarrow$zh}
                \addplot[
                    color=blue,
                    dotted,
                    mark options={solid},
                    mark=*,
                    line width=1.5pt,
                    mark size=2pt
                    ]
                    coordinates {
                    (0, 52.9)
                    (64, 54.8)
                    (128, 55.6)
                    (192, 55.7)
                    (256, 55.7)
                    };
                    \addlegendentry{en$\rightarrow$tr}
                \end{axis}
                \end{tikzpicture}
            %\label{fig:flops_intro}
    }
    \subfigure[Performance vs. $K_{r}$]{\label{fig:parameters_kr}
            \begin{tikzpicture}[font=\Large, scale=0.45]
                \begin{axis}[
                    legend cell align={left},
                    legend style={nodes={scale=1.0, transform shape}},
                    xlabel={\# of random negatives $K_r$},
                    xtick pos=left,
                    tick label style={font=\large},
                    %ylabel style={font=\large},
                    %xmode=log,
                    %log ticks with fixed point,
                    %ylabel={p1},
                    %xmin=0.005, xmax=100.0,
                    %ymin=0.33, ymax=0.6,
                    xtick={ 0, 64, 128, 192, 256},
                    xticklabels={ $0$, $64$, $128$, $192$, $256$},
                    ytick={49.0, 50.0, 51.0,52.0,53.0,54.0,55.0,56.0},
                    yticklabels={$49.0$, $50.0$, $51.0$, $52.0$, $53.0$, $54.0$, $55.0$, $56.0$},
                    legend pos=south east,
                    ymajorgrids=true,
                    grid style=dashed
                ]
                \addplot[
                    color=purple,
                    dotted,
                    mark options={solid},
                    mark=diamond*,
                    line width=1.5pt,
                    mark size=2pt
                    ]
                    coordinates {
                    (0, 48.5)
                    (64, 51.2)
                    (128, 51.8)
                    (192, 51.7)
                    (256, 51.9)
                    };
                    \addlegendentry{en$\rightarrow$zh}
                \addplot[
                    color=blue,
                    dotted,
                    mark options={solid},
                    mark=*,
                    line width=1.5pt,
                    mark size=2pt
                    ]
                    coordinates {
                    (0, 53.3)
                    (64, 55.1)
                    (128, 55.5)
                    (192, 55.7)
                    (256, 55.7)
                    };
                    \addlegendentry{en$\rightarrow$tr}
                \end{axis}
                \end{tikzpicture}
            %\label{fig:flops_intro}
    }
\subfigure[Performance vs. $n$]{\label{fig:parameters_n}
            \begin{tikzpicture}[font=\Large, scale=0.45]
                \begin{axis}[
                    legend cell align={left},
                    legend style={nodes={scale=1.0, transform shape}},
                    xlabel={Number of HMs $n$},
                    xtick pos=left,
                    tick label style={font=\large},
                    %ylabel style={font=\large},
                    %xmode=log,
                    %log ticks with fixed point,
                    %ylabel={p1},
                    %xmin=0.005, xmax=100.0,
                    %ymin=0.33, ymax=0.6,
                    xtick={200, 250, 300, 350, 400},
                    xticklabels={$200$, $250$, $300$, $350$, $400$},
                    ytick={49.0, 50.0, 51.0,52.0,53.0,54.0,55.0,56.0},
                    yticklabels={$49.0$, $50.0$, $51.0$, $52.0$, $53.0$, $54.0$, $55.0$, $56.0$},
                    legend pos=south east,
                    ymajorgrids=true,
                    grid style=dashed
                ]
                \addplot[
                    color=purple,
                    dotted,
                    mark options={solid},
                    mark=diamond*,
                    line width=1.5pt,
                    mark size=2pt
                    ]
                    coordinates {
                    (200, 49.9)
                    (250, 50.6)
                    (300, 51.9)
                    (350, 51.7)
                    (400, 51.8)
                    };
                    \addlegendentry{en$\rightarrow$zh}
                \addplot[
                    color=blue,
                    dotted,
                    mark options={solid},
                    mark=*,
                    line width=1.5pt,
                    mark size=2pt
                    ]
                    coordinates {
                    (200, 53.7)
                    (250, 55.1)
                    (300, 55.7)
                    (350, 55.6)
                    (400, 55.5)
                    };
                    \addlegendentry{en$\rightarrow$tr}
                \end{axis}
                \end{tikzpicture}
            %\label{fig:flops_intro}
    }
    \subfigure[Performance vs. $\tau_s$]{
            \label{fig:parameters_s}
            \begin{tikzpicture}[font=\Large, scale=0.45]
                \begin{axis}[
                    legend cell align={left},
                    legend style={nodes={scale=1.0, transform shape}},
                    xlabel={The adapter threshold $\tau_s$},
                    xtick pos=left,
                    tick label style={font=\large},
                    %ylabel style={font=\large},
                    %xmode=log,
                    %log ticks with fixed point,
                    %ylabel={p1},
                    %xmin=0.005, xmax=100.0,
                    %ymin=0.33, ymax=0.6,
                    xtick={ 0.5, 0.6, 0.7, 0.8, 0.9},
                    xticklabels={ $0.5$,$0.6$,$0.7$, $0.8$, $0.9$},
                    ytick={49.0, 50.0, 51.0,52.0,53.0,54.0,55.0,56.0},
                    yticklabels={$49.0$, $50.0$, $51.0$, $52.0$, $53.0$, $54.0$, $55.0$, $56.0$},
                    legend pos=south east,
                    ymajorgrids=true,
                    grid style=dashed
                ]
                \addplot[
                    color=purple,
                    dotted,
                    mark options={solid},
                    mark=diamond*,
                    line width=1.5pt,
                    mark size=2pt
                    ]
                    coordinates {
                    (0.5, 48.6)
                    (0.55, 49.0)
                    (0.6, 49.3)
                    (0.65, 49.9)
                    (0.7, 50.3)
                    (0.75, 50.6)
                    (0.8, 50.9)
                    (0.85, 51.3)
                    (0.9, 51.6)
                    (0.95, 51.4)
                    };
                    \addlegendentry{en$\rightarrow$zh}
                \addplot[
                    color=blue,
                    dotted,
                    mark options={solid},
                    mark=*,
                    line width=1.5pt,
                    mark size=2pt
                    ]
                    coordinates {
                    (0.5, 52.3)
                    (0.55, 52.1)
                    (0.6, 52.7)
                    (0.65, 53.3)
                    (0.7, 54.0)
                    (0.75, 54.7)
                    (0.8, 55.1)
                    (0.85, 55.7)
                    (0.9, 55.5)
                    (0.95, 55.0)
                    };
                    \addlegendentry{en$\rightarrow$tr}
                \end{axis}
                \end{tikzpicture}
            %\label{fig:flops_intro}
    }    
    \caption{Parameter sensitivity analysis.} 
\end{figure}

\paragraph{Number of negative samples} 
Figure \ref{fig:parameters_kh} and Figure \ref{fig:parameters_kr} demonstrate the performance curves of the number of hard negative samples $K_{h}$ and random samples $K_{r}$, respectively.   
One can clearly see that both types of negative sampling strategies contribute to improving model performance, which indicates that each strategy provides its unique crucial signals to guide the optimization process. 
% \czl{In addition, the hard negative sampling is more important than the random samples because the performance of $K_{h}=0$ is significant inferior to $K_{r}=0$. 
% This is reasonable as the hard negative samples are generated based on the current model checkpoint, which are more difficult to be distinguished and empower RAPO with discriminative ability. } 

\paragraph{Number of Householder matrices} As shown in Figure \ref{fig:parameters_n}, with the increases of $n$, the performance of RAPO first increases and then keeps steady. This is reasonable as more Householder matrices means the projection could be conducted in a larger rotation space, which benefits the induction performance at the beginning. 
When $n$ is equal to or larger than the dimension of word embeddings $d=300$, the Householder projection reaches the peak of its expressivity and thus cannot bringing more benefits. 

\paragraph{Threshold of adaptor} 
The hyper-parameter $\tau_{s}$ denotes the threshold to select the neighbor words to generate the contextual semantic vectors. 
As shown in Figure \ref{fig:parameters_s}, model performance first increases and then significant declines when $\tau_{s}$ is enlarged from 0.5 to 0.95. 
On one hand, a small $\tau_{s}$ incorporates more neighbor words to provide richer semantics, while also aggravates the risk of introducing noises. 
On the other hand, a large $\tau_{s}$ ensures the reliability of the contextual vectors but may lead to the scarce neighborhood information.  
Thus,  $\tau_{s}$ should be carefully tuned to find the balance between the contextual semantics and potential noises.

\section{Conclusion}
In this paper, we propose a novel model RAPO for bilingual lexicon induction. 
Different from previous works, RAPO is formulated as a ranking paradigm, which is more suitable to the nature of studied tasks.  
Two novel modules are further employed by deeply mining the unique characteristics of BLI task: the Householder projection to ensure the strict orthogonal mapping function, and the personalized adapter to learn unique embedding offsets for different words. 
Extensive experimental results on 20 datasets demonstrate the superiority of our proposal.

\section*{Limitations}
Despite the promising performance of the proposed RAPO model, it may suffer from the following limitations:

\begin{itemize}
	\item RAPO has more hyper-parameters than the previous works, leading to the exhausting and time-consuming hyper-parameter tuning process. 
	
	\item Though RAPO has achieved the best performance over almost all the datasets, it fails in  few datasets such as en$\rightarrow$tr, which might be caused by the insufficient hyper-parameter tuning.   
	
	\item The supervised signals are indispensable to our proposal, and thus RAPO cannot be applied to the unsupervised learning setting without any labeled data. 
\end{itemize}

\bibliography{anthology,custom}
\bibliographystyle{acl_natbib}

\clearpage

\appendix

\section{Notations}
\label{sec:notations}

For the sake of clarification, notations used in this paper are listed in Table \ref{params} .

\begin{table*}[h]
\begin{center}
\begin{small}
\renewcommand{\arraystretch}{1.5}
\begin{tabular}{ccl}
\toprule
Symbol     & Shape & Description  \\
\midrule
$n_x(n_y)$ & $\mathbb{R}$ & Vocabulary size of source(target) language \\
$d$ & $\mathbb{R}$ & Embedding size \\
$\mb{X}$ & $\mathbb{R} ^ {d \times n_x}$ & Monolingual embedding matrices consisting of $n_x$ words for source language  \\
$\mb{Y}$ & $\mathbb{R} ^ {d \times n_y}$ & Monolingual embedding matrices consisting of $n_y$ words for target language  \\
$\mathcal{D}$ & $-$ & Set of available aligned seed dictionary  \\
$l$ & $\mathbb{R}$ & Size of $\mathcal{D}$  \\
$\phi_s$($\phi_t$) & $-$ & Mapping function for source(target) language  \\
$\bar{x}$($\bar{y}$) & $\mathbb{R} ^ {d}$ & Contextual semantic vector of source word $\mb{x}$ ( target word $\mb{y}$ )  \\
$\tau_s$($\tau_t$) & $\mathbb{R}$ & Similarity threshold of personalized adapter for source(target) language \\
$\mathcal{M}_s$($\mathcal{M}_t$) & $-$ & Set of contextual semantic neighbor words in the source(target) space \\
$m_s$($m_t$) & $\mathbb{R}$ & Size of $\mathcal{M}_s$($\mathcal{M}_t$) \\
$W_s$($W_t$)  & $\mathbb{R} ^ {d \times d}$  & Learnable parameters of personalized adapter for source(target) language \\
$\sigma$  & $ - $  & Activation function used in personalized adapter\\
$\tilde{x}$($\tilde{y}$) & $\mathbb{R} ^ {d}$ & Calibrated embedding of source word $\mb{x}$ (target word $\mb{y}$)  \\
$\mathcal{V}_s$($\mathcal{U}_t$)     &  $-$       &  Set of unit vectors to parameterize the Householder projection for the source(target) language \\
$n$     &  $\mathbb{R}$      &  Number of unit vectors to parameterize the Householder projection \\
$\hat{x}$($\hat{y}$) & $\mathbb{R} ^ {d}$ & Mapped embedding in the shared latent space of source word embedding $\mb{x}$ (target word $\mb{y}$) \\
$\mathcal{N}^{-}$ & $-$ & Set of negative samples  \\
$K$ & $\mathbb{R}$  & Number of negative samples  \\
$K_h$ & $\mathbb{R}$  & Number of dynamic hard negative samples  \\
$K_r$ & $\mathbb{R}$  & Number of random negative samples  \\
$\theta$ & $-$ & Parameters of RAPO, including $\{\mb{W}_s$, $\mb{W}_t$, $\mathcal{V}_s$, $\mathcal{U}_t\}$ \\

\bottomrule
\end{tabular}
\end{small}
\end{center}
\caption{Notations used in this paper.}
\label{params}
\end{table*}

\section{Proofs of Theorem \ref{thm-1}}
\label{proof}

\begin{table*}[t]
\small
  \centering
  %\resizebox{1\columnwidth}{!}{
  \setlength{\tabcolsep}{2mm}{
    \begin{tabular}{l|cc|cc|cc|cc|cc}
      \hline
     \multirow{2}{*}{Statics} & \multicolumn{2}{c|}{en-es}  & \multicolumn{2}{c|}{en-fr} & \multicolumn{2}{c|}{en-it}  & \multicolumn{2}{c|}{en-ru}  & \multicolumn{2}{c}{en-zh} \\
       &$\rightarrow$ &  $\leftarrow$ &$\rightarrow$ & $\leftarrow$ &$\rightarrow$ & $\leftarrow$ &$\rightarrow$ & $\leftarrow$ &$\rightarrow$ & $\leftarrow$ \\
      \hline
      \# train set    & 11977 & 8667 & 10872 & 8270 & 9657 & 7364 & 10887 & 7452 & 8728 & 8891\\
      \# test set   &2975 & 2416 & 2943 & 2342 & 2585 & 2102 & 2447 & 2069 & 2230 & 2483\\
      \# unique source words   &6500 & 6500 & 6500 & 6500 & 6500 & 6500 & 6500 & 6500 & 6500 & 6500\\
      \# unique target words     &12411 & 8373 & 11524 & 8309 & 10762 & 7562 & 11491 & 6811 & 8102 & 8618\\
    %   \multicolumn{13}{c}{Semi-Supervised use "all 5k"}      \\
    %   \hline
    %   BLISS(R) & 84.3 & - & 79.1 & - & 83.9 & - & 79.3   &  -   & 57.1 & 67.7 & 48.7 & 47.3 \\
    %   LNMAP(linear/non-linear) \\
    %   CSS/PSS RCSLS  \\
       \hline
    \end{tabular}
  }
  \caption{Statistics of rich-resource language pairs in MUSE. }
  \label{tab:datasets_rich}
\end{table*}

\begin{table*}[t]
\small
  \centering
  %\resizebox{1\columnwidth}{!}{
  \setlength{\tabcolsep}{2mm}{
    \begin{tabular}{l|cc|cc|cc|cc|cc}
      \hline
     \multirow{2}{*}{Statics} & \multicolumn{2}{c|}{en-fa}  & \multicolumn{2}{c|}{en-tr} & \multicolumn{2}{c|}{en-he}  & \multicolumn{2}{c|}{en-ar}  & \multicolumn{2}{c}{en-et} \\
       &$\rightarrow$ &  $\leftarrow$ &$\rightarrow$ & $\leftarrow$ &$\rightarrow$ & $\leftarrow$ &$\rightarrow$ & $\leftarrow$ &$\rightarrow$ & $\leftarrow$ \\
      \hline
      \# train set    & 8869 & 8510 & 9771 & 8793 & 9634 & 7737 & 11571 & 7534 & 8261 & 6509 \\
      \# test set   &2148 & 2202 & 2261 & 2291 & 2379 & 2195 & 2695 & 2061 & 1991 & 1817\\
      \# unique source words   &6498 & 6451 & 6498 & 6442 & 6495 & 6462 & 6500 & 6453 & 6500 & 6500\\
      \# unique target words     &8837 & 8691 & 9259 & 8589 & 10018 & 7458 & 11787 & 6144 & 8991 & 6796\\
    %   \multicolumn{13}{c}{Semi-Supervised use "all 5k"}      \\
    %   \hline
    %   BLISS(R) & 84.3 & - & 79.1 & - & 83.9 & - & 79.3   &  -   & 57.1 & 67.7 & 48.7 & 47.3 \\
    %   LNMAP(linear/non-linear) \\
    %   CSS/PSS RCSLS  \\
       \hline
    \end{tabular}
  }
  \caption{Statistics of low-resource language pairs in MUSE. }
  \label{tab:datasets_low}
\end{table*}

Householder projection is composed of a set of consecutive Householder matrices. 
Specifically, given a series of unit vectors $\mathcal{V} =  \{\mb{v}_{i}\}_{i=1}^{n}$ where $\mb{v}_i\in \mathbb{R}^d$ and $n$ is a positive integer, we define the Householder Projection (HP) as follows:
\begin{equation}
	% \vspace{-1mm}
	\begin{split}
		\label{map-Hrot-proof}
		{\rm HP}(\mathcal{V}) = \prod_{i=1}^{n}H(\mb{v}_{i}).
	\end{split}
	% \vspace{-1mm}
\end{equation}

% Any $k \times k$ orthogonal matrix $Q$ can be decomposed into the product of $k-1$ or $k$ Householder matrices.
The image of ${\rm HP}$ is the set of all $n \times n$ orthogonal matrices, i.e., ${\rm Image(HP)}=\boldsymbol{{\rm O}}(n)$, $\boldsymbol{{\rm O}}(n)$ is the $n$-dimensional orthogonal group.

We first prove that the image of ${\rm HP}$ is a subset of $\boldsymbol{{\rm O}}(n)$. Note that each Householder matrix is orthogonal. Therefore, the product of $n$ Householder matrices is also an orthogonal matrix, i.e., ${\rm Image(HP)} \subset \boldsymbol{{\rm O}}(n)$.

Then we also prove that its converse is also valid, i.e., any $n \times n$ orthogonal matrix $Q$ can be decomposed into the product of $n$ Householder matrices.
From the Householder QR decomposition \cite{householder1958unitary}, we can upper triangularize any full-rank matrix $W \in \mathbb{R}^{n \times n}$ by using $n-1$ Householder matrices, i.e.,
\begin{equation}
	H(\mb{v}_{n-1})H(\mb{v}_{n-2})\cdots H(\mb{v}_1)W=R,\nonumber
\end{equation}
where $R\in \mathbb{R}^{n \times n}$ is an upper triangular matrix and its first $n-1$ diagonal elements are all positive. 
%the unit vectors $u_i$ that define each $H({u_i})$ are routinely computed from $U$.

When Household QR decomposition is performed on an orthogonal matrix $Q$, we can get: 
\begin{equation}
	H(\mb{v}_{n-1})H(\mb{v}_{n-2})\cdots H(\mb{v}_1)Q=R.\nonumber
\end{equation}
Note that $R$ here is both upper triangular and orthogonal (i.e., $RR^T=I$) since it is a product of $n$ orthogonal matrices. 
It establishes that $R$ is a diagonal matrix, where the first $n-1$ diagonal entries are equal to $+1$ and the last diagonal entry is either $+1$ or $-1$. 

% the diagonal entry of $-1$ can only appear in the last column.

If the last diagonal entry of $R$ is equal to $+1$, i.e., $R=I$, we can set $\mb{v}_n=(0,\ldots,0,0)^\top \in \mathbb{R}^{n}$ and consequently get
\begin{equation}
	H(\mb{v}_{n-1})H(\mb{v}_{n-2})\cdots H(\mb{v}_1)Q=I=H(\mb{v}_n).\nonumber
\end{equation}
As each Householder matrix $H(\mb{v}_i)$ is its own inverse, we obtain that
\begin{equation}
	\label{k-1}
	Q=H(\mb{v}_1)\cdots H(\mb{v}_{n-1})H(\mb{v}_{n}).\nonumber
\end{equation}

If the last diagonal entry of $R$ is equal to $-1$, we can set $\mb{v}_n=(0,\ldots,0,1)^\top \in \mathbb{R}^{n}$ and get the same conclusion.

% \begin{equation}
%     H(u_k)R=H(u_k)H(u_{k-1})\cdots H(u_1)Q=I.\nonumber
% \end{equation}
% Since $H(u_i)$ is its own inverse, we also obtain that
% \begin{equation}
% \label{k}
%     Q=H(u_1)\cdots H(u_{k-1})H(u_{k}).
% \end{equation}
%As each Householder matrix $H_{u_i}$ is its own inverse, we obtain that $Q=H_{u_1}\cdots H_{u_{k-1}}H_{u_{k}}$.
From the above we can see that any $n \times n$ orthogonal matrix can be decomposed into the product of $n$ Householder matrices, i.e., $\boldsymbol{{\rm O}}(n) \subset {\rm Image(HP)}$. All in all, we have ${\rm Image(HP)} = \boldsymbol{{\rm O}}(n)$.

\section{Complexity analysis}
\label{sec:Complexity}

\paragraph{Parameter complexity analysis} The learnable parameters in RAPO come from two modules:  personalized offset adapter and Householder projections. 
The learnable parameters in adapter is matrix $W_s$ with the shape of $d^2$. 
The learnable parameters in Householder transformation include $n$ $d$-dimension unit vectors $\mathcal{V}_s = \{\mb{v}_1,..,\mb{v}_n\}$, which has a total size of $nd$. Therefore, the number of parameters for each language is $O(d^2+nd)$. In our experiments, the number of householder reflection $n$ is the same to $d$. Thus, the final parameter complexity of RAPO is $O(d^2)$, which is same to the previous models .

\paragraph{Time complexity analysis} Given a word embedding $\mb{x}$ of dimension $d$, we first need to find its contextual semantic vector $\bar{\mb{x}}$ of dimension $d$, which can be computed in advance and stored in an efficient lookup table. Next, we calibrate embedding $\tilde{\mb{x}}$ using the personalized adapter according to Formula (\ref{adapter_calc}), which has a time complexity of $O(d^2)$. Then, the calibrated embedding $\tilde{\mb{x}}$ will be mapped to the shared latent space through Householder projection based on the Formula (\ref{HT}). The time complexity of Formula (\ref{HT}) is $O(nd^2)$, in which $n$ matrix-vector multiplications incur high computational costs.
However, it is worth noting that these matrix multiplications can be replaced by the vector operations. 
Formally, based on Equation (\ref{HT_vec}), the $j$-th matrix-vector multiplication can be expressed as:
\begin{equation}
\label{iter-eff}
    \begin{split}
        \mb{\hat{x}}_j &= \operatorname{H}(\mb{v}_j)\mb{\hat{x}}_{j-1} \\
        &=\mb{\hat{x}}_{j-1} - 2\braket{\mb{\hat{x}}_{j-1}, \mb{v}_j}\mb{v}_j
    \end{split}
\end{equation}
%as shown in Equation (\ref{ref vector operation}).
where $\mb{\hat{x}}_0 =\tilde{\mb{x}}$. 
Through such iterated vector operations, the time complexity can be reduced to $O(nd)$ or $O(d^2)$. Therefore, the time complexity of mapping function in RAPO is $O(2d^2)$, which is linearly comparable with previous methods.

% \subsection{Discussion} 

% \subsubsection{Connections to previous works}

% \subsubsection{Complexity Analysis} 

% \subsubsection{Non-iscomo lanagueg handeling}
% We use the Faiss tools to retrieve nearest neighbors. 

\section{Dataset statistics}
\label{sec:datasets}

The widely used MUSE dataset \citep{DBLP:conf/iclr/LampleCRDJ18} consists of FastText monolingual embeddings of 300 dimensions \citep{bojanowski-etal-2017-enriching} trained on Wikipedia monolingual corpus and gold dictionaries for many language pairs divided into training and test sets.
We conduct extensive experiments over multiple language pairs in the MUSE dataset, including five popular rich-resource language pairs (French (fr), Spanish(es), Italian (it), Russian (ru), Chinese (zh) from and to English(en)) and five low-resource language pairs (Faroese(fa), Turkish(tr), Hebrew(he), Arabic(ar), Estonian(er) from and to English(en)), totally 20 BLI datasets considering bidirectional translation. Table \ref{tab:datasets_rich} and \ref{tab:datasets_low} summarizes the detailed statistics of rich- and low-resource language pairs, respectively.

\begin{table}[h]
\begin{center}
\begin{small}
\begin{tabular}{ccc}
\toprule
Hyperparameter  & Search Space & Type  \\
\midrule
$K_h$  & $\{64, 128, 192, 256\}$ & Choice\\
$K_r$  & $\{64, 128, 192, 256\}$ & Choice\\
$\sigma$  & $\{$ none, tanh, sigmoid $\}$ & Choice\\
$lr$  & $[0.001, 0.003]$ & Range\\
$\tau_s$ & $[0.7, 0.99]$ & Range\\
$\tau_t$  & $[0.7, 0.99]$ & Range\\
$\lambda_1$  & $[0.5, 2.5]$ & Range\\
$\lambda_2$  & $[0.001, 0.1]$ & Range\\
\bottomrule
\end{tabular}
\end{small}
\end{center}
\caption{Hyper-parameter search space.}
\label{Hyper_search_tab}
\end{table}

\section{Related Work}
\label{relatedwork}

Recent proposed work on BLI can be mainly divided into two categories. The first is unsupervised learning methods, which induces dictionaries according to the characteristics of the embedding space. The second is the supervised and semi-supervised learning methods, which trains the model based on a small seed dictionary.

\paragraph{Unsupervised method} The major challenge of unsupervised methods is to build high-quality initial seed dictionary.
\citet{DBLP:conf/iclr/LampleCRDJ18} are the first to show impressive results for unsupervised word translation by pairing adversarial training with effective refinement methods. Given two monolingual word embeddings, they proposed an adversarial training method, where a linear mapper (generator) plays against a discriminator. They also impose the orthogonality constraint on the mapper. After adversarial training, they use the iterative Procrustes solution similar to their supervised approach. \citet{artetxe-etal-2018-robust} learn an initial dictionary by exploiting the structural similarity of the embeddings in an unsupervised way. They propose a robust self-learning to improve it iteratively.\citet{mohiuddin-joty-2019-revisiting} use adversarial autoencoder for unsupervised word translation. They use linear autoencoders in their model, and the mappers are also linear. \citet{ren-etal-2020-graph} propose a graph-based paradigm to induce bilingual lexicons. They first build a graph for each language with its vertices representing different words. Then they extract word cliques from the graphs and align the cliques of two languages to induce the initial word translation solution. Their methods achieved SOTA results in the unsupervised setting.

\paragraph{Supervised and Semi-supervised methods}
Supervised and semi-supervised methods mainly focus on learning more efficient mapping function to improve the induction performance. \citet{DBLP:conf/aaai/ArtetxeLA18} propose a multi-step framework based on their unsupervised method. Their framework consists of several steps: whitening, orthogonal mapping, re-weighting, de-whitening, and dimensionality reduction. \citet{joulin-etal-2018-loss} proposed to directly include the CSLS criterion in the learning object and maximum the CSLS score between the translation pairs of seed dictionary in order to make learning and inference consistent. \citet{jawanpuria-etal-2020-geometry} proposed to map the source and target words from their original embedding spaces to a common latent space via language-specific orthogonal transformations. They further define a learnable Mahalanobis similarity metric, which allows for a more effective similarity comparison of embeddings. \citet{sogaard-etal-2018-limitations} empirically show that even closely related languages are far from being isomorphic and \citet{patra-etal-2019-bilingual} propose a semi-supervised technique that relaxes the isomorphic assumption while leveraging both seed dictionary pairs and a larger set of unaligned word embeddings. \citet{mohiuddin-etal-2020-lnmap} uses non-linear mapping in the latent space of two independently pre-trained autoencoders, which is also independent of the isomorphic assumption. \citet{zhao-etal-2020-semi} design two message-passing mechanisms in semi-supervised setting to transfer knowledge between annotated and non-annotated data, named prior optimal transport and bi-directional lexicon update respectively. \citet{glavas-vulic-2020-non} proposed to move each point along an instance-specific translation vector estimated from the translation vectors of nearest neighbours in training dictionary. Notably, comparing with these supervised and semi-supervised methods, our RAPO model aims at optimizing ranking-base objectives , which is more suitable to the induction task. Furthermore, with proposed personalized adapter and Householder projection, RAPO enjoys the merits from the unique traits of each word and the global consistency across languages, which is capable of 
improving induction performance on different language pairs consistently.

\section{Hyper-parameter search}
\label{sec:hypter_search}
The hyper-parameters are tuned by the random search \citep{bergstra2012random} for each BLI dataset, including number of dynamic hard negative samples $K_h$, number of negative samples $K_r$, activation function used in personalized adapter $\sigma$, learning rate $lr$, similarity threshold of personalized adapter for source and target language $\tau_s$,$\tau_t$, and the weights in loss function $\lambda_1, \lambda_2$. The hyper-parameter search space is shown in Table \ref{Hyper_search_tab}.

\end{document}